\documentclass{article} 
\usepackage{iclr2023_conference,times}

\usepackage{amsmath,amsfonts,bm}

\def\eqref#1{equation~\ref{#1}}

\def\1{\bm{1}}

\DeclareMathAlphabet{\mathsfit}{\encodingdefault}{\sfdefault}{m}{sl}
\SetMathAlphabet{\mathsfit}{bold}{\encodingdefault}{\sfdefault}{bx}{n}

\usepackage{hyperref}
\usepackage{url}
\iclrfinalcopy

\usepackage[pdftex]{graphicx}
\usepackage{algorithm, algorithmic}
\usepackage{multicol, multirow}
\def\ourmethod{REVD }
\def\renyi{R\'enyi }
\def\ie{{\em i.e.}}

\title{Rewarding Episodic Visitation Discrepancy for Exploration in Reinforcement Learning}

\author{
Mingqi Yuan\textsuperscript{\textbf{1}}, Bo Li\textsuperscript{\textbf{1}}, Xin Jin\textsuperscript{\textbf{2}} \& Wenjun Zeng\textsuperscript{\textbf{2}} \\
\textsuperscript{\textbf{1}}Department of Computing, The Hong Kong Polytechnic University\\
\textsuperscript{\textbf{2}}Eastern Institute for Advanced Study
}

\begin{document}

\maketitle

\begin{abstract}
Exploration is critical for deep reinforcement learning in complex environments with high-dimensional observations and sparse rewards. To address this problem, recent approaches propose to leverage intrinsic rewards to improve exploration, such as novelty-based exploration and prediction-based exploration. However, many intrinsic reward modules require sophisticated structures and representation learning, resulting in prohibitive computational complexity and unstable performance. In this paper, we propose Rewarding Episodic Visitation Discrepancy (REVD), a computation-efficient and quantified exploration method. More specifically, REVD provides intrinsic rewards by evaluating the \renyi divergence-based visitation discrepancy between episodes. To estimate the divergence efficiently, a $k$-nearest neighbor estimator is utilized with a randomly-initialized state encoder. Finally, the REVD is tested on Atari games and PyBullet Robotics Environments. Extensive experiments demonstrate that REVD can significantly improve the sample efficiency of reinforcement learning algorithms and outperform the benchmarking methods.
\end{abstract}

\section{Introduction}
Efficient and complete exploration remains one of the significant challenges in deep reinforcement learning. Inspired by human learning patterns, recent approaches propose to leverage intrinsic motivation to improve exploration, especially when the agent receives sparse or missing rewards in real-world scenarios \citep{schmidhuber1991possibility, oudeyer2008can, oudeyer2007intrinsic}. Intrinsic rewards encourage the agent to visit novel states \citep{bellemare2016unifying, ostrovski2017count, tang2017exploration, burda2018exploration} or to increase its knowledge and prediction accuracy of the environment dynamics \citep{stadie2015incentivizing, pathak2017curiosity, yu2020intrinsic, houthooft2016vime, oh2015action}. For instance, \citet{ostrovski2017count, bellemare2016unifying} leveraged a density model to evaluate the novelty of states, and the intrinsic rewards are defined to be inversely proportional to the visiting frequencies. As a result, the agent is motivated to revisit the infrequently-seen states and realize comprehensive exploration. In contrast, \cite{stadie2015incentivizing, pathak2017curiosity, yu2020intrinsic} designed an auxiliary model to learn the dynamics of the environment, and the prediction error is utilized as the intrinsic reward. Furthermore, \cite{burda2018large} performed experiments on Atari games only using the prediction-based intrinsic rewards, demonstrating that the agent could also achieve remarkable performance.

However, the aforementioned methods cannot provide sustainable exploration incentives because the exploration bonus will vanish as the agent gets to know the environment. As a result, the agent may miss the downstream learning opportunities and fail to learn in hard-exploration tasks. To address the problem, \citet{badia2020never} proposed a never-give-up (NGU) framework that produces intrinsic rewards composed of episodic and long-term state novelty. The episodic state novelty is evaluated using an episodic memory and pseudo-count method, encouraging the agent to visit diverse states in each episode. Since the memory is erased at the beginning of each episode, the intrinsic reward will not vanish across the episodes. In contrast, \citet{raileanu2020ride} proposed a more straightforward method entitled rewarding-impact-driven-exploration (RIDE). RIDE takes the difference between two consecutive encoded states as the intrinsic reward, encouraging the agent to choose actions that result in large state changes. However, NGU and RIDE are expensive and unstable because they rely heavily on auxiliary models. Poor representation learning may incur severe performance loss, making it difficult to be applied to arbitrary tasks. 

In this paper, we propose a novel intrinsic reward method entitled \textbf{R}ewarding \textbf{E}pisodic \textbf{V}isitation \textbf{D}iscrepancy (REVD), which provides powerful exploration incentives with simple, computation-efficient, and robust architecture. Our main contributions can be summarized as follows:
\begin{itemize}
	\item REVD requires no memory model or database, which overcomes the problem of vanishing intrinsic rewards and provides sustainable exploration bonuses;
	\item REVD significantly promotes the sample efficiency of reinforcement learning algorithms without introducing any representation learning and additional models;
	\item REVD provides a quantified and intuitive metric to evaluate the exploration change between episodes;
	\item Extensive experiments are performed to compare the performance of REVD against existing methods using Atari games and PyBullet Robotics Environments. Simulation results demonstrate that REVD achieves remarkable performance and outperforms the benchmarking schemes.
\end{itemize}
 
\begin{figure}
	\centering
	\includegraphics[width=0.9\linewidth]{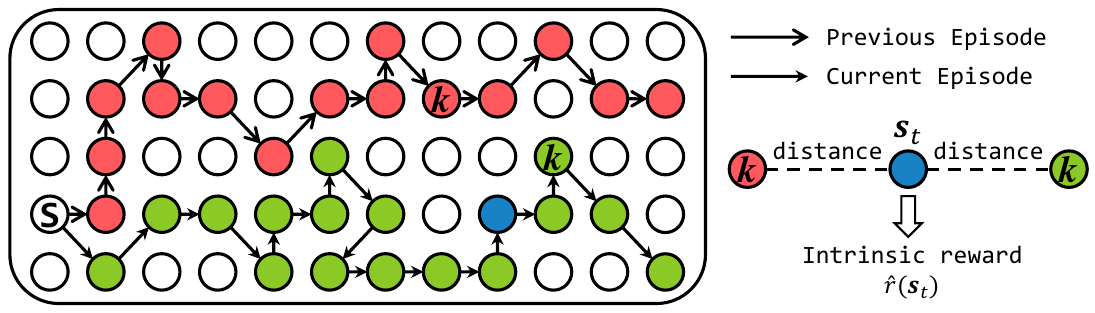}
	\caption{The overview of REVD. Here, \textbf{S} represents the starting point, node with a label $\bm{k}$ represents the $k$-nearest neighbor ($k$-NN) of $\bm{s}_{t}$ among the previous episode and the current episode. REVD computes the intrinsic reward using the two distances between $\bm{s}_{t}$ and its $k$-NN states.}
	\label{fig:revd}
\end{figure}

\section{Related Work}
\subsection{Novelty-Based Exploration}
Novelty-based exploration methods encourage the agent to explore novel states or state-action pairs. \citet{strehl2008analysis} utilized the state visitation counts as the exploration bonus in tabular setting and provided theoretical justification for its effectiveness. \citet{bellemare2016unifying, machado2020count, ijcai2017p0344, bellemare2016unifying, ostrovski2017count, tang2017exploration, burda2018exploration} extended this method to environments with high-dimensional observations. More specifically, \citet{burda2018exploration} employed a predictor network and target network (fixed and randomly-initialized) to record the visited states and allocated higher rewards to infrequently-seen states. \citet{ijcai2017p0344} proposed a $\phi$-pseudo count method by exploiting the feature space of value function approximation, which can estimate the uncertainty related to any state. \citet{machado2020count} designed the exploration bonus using the norm of the successor representation of states, and improved the exploration by capturing the diffusion properties of the environment. However, these novelty-based approaches suffer from vanishing intrinsic rewards and cannot maintain exploration across episodes. Our method can effectively address this problem because it requires no additional memory models or database.

\subsection{Prediction-Based Exploration}
Prediction-based exploration methods encourage the agent to take actions that reduce the uncertainty of predicting the consequence of its own actions \citep{stadie2015incentivizing, pathak2017curiosity, burda2018large, yu2020intrinsic}. \citet{pathak2017curiosity} proposed an intrinsic-curiosity-module (ICM) framework, which uses an inverse-forward pattern to learn a representation space of state and defined the intrinsic reward as the error of predicting the encoded next-state. A similar representation learning technique is also leveraged in RIDE \citep{raileanu2020ride}. In particular, \citet{yu2020intrinsic} introduced the intrinsic reward concept to imitation learning and redesigned the ICM using the variational auto-encoder \citep{kingma2013auto}. However, prediction-based approaches also suffer from vanishing intrinsic rewards and sophisticated representation learning, making it expensive and unstable. Our method follows \citep{seo2021state} and takes a randomly-initialized neural network to encode the state space, which is cheap and robust to be applied to arbitrary tasks.

\subsection{Computation-Efficient Exploration}
Computation-efficient exploration aims to provide intrinsic rewards in a cheap, robust, and generalized pattern \citep{seo2021state, mutti2021task, yuan2022renyi}. \citet{seo2021state} suggested maximizing the state entropy and designed a random-encoder-for-efficient-exploration (RE3) method, which uses a $k$-NN estimator to make efficient entropy estimation and divides the sample mean into particle-based intrinsic rewards. RE3 can significantly promote the sample efficiency of model-based and model-free reinforcement learning algorithms without any representation learning. However, \citet{zhang2021exploration} indicated that maximizing the Shannon entropy may lead to a policy that visits some state-action pairs with a vanishing probability and proposed to maximize the \renyi entropy of the state-action distribution (MaxR\'enyi). \citet{yuan2022renyi} combined the RE3 and Max\renyi by introducing the $k$-NN estimator and proposed a \renyi state entropy maximization method, and discussed $k$-value selection and state space encoding in detail. However, \citet{poczos2011estimation} found that the $k$-NN estimators employed in their work may lead to severe estimation errors because of the inappropriate use of the reverse Fatou lemma. In this paper, we leverage an improved $k$-NN estimator proposed by \citep{poczos2011estimation} to make an efficient divergence estimation. Our method also provides an intuitive and quantified metric to reflect the exploration change between episodes. A closely related work to us is MADE proposed by \citep{zhang2021made} that maximizes the deviation from the explored regions. However, MADE needs to estimate the visitation density over state-action pairs, which is non-trivial in complex environments with high-dimensional observations. Therefore, our method is more efficient and can be easily deployed in arbitrary tasks.

\section{Preliminaries}
\subsection{Problem Formulation}
In this paper, we study the reinforcement learning problem considering a single agent Markov decision process (MDP), which can be defined as a tuple $\mathcal{M}=\langle\mathcal{S},\mathcal{A},P,\check{r},\varsigma(\bm{s}_0),\gamma\rangle$ \citep{bellman1957markovian}. Here, $\mathcal{S}$ is the state space, $\mathcal{A}$ is the action space, $P:\mathcal{S}\times\mathcal{A}\rightarrow\Pi(\mathcal{S})$ is the state-transition function, where $\Pi(\mathcal{S})$ is the space of probability distributions over $\mathcal{S}$. In addition, $\check{r}:\mathcal{S}\times\mathcal{A}\rightarrow\mathbb{R}$ is the extrinsic reward function, $\varsigma(\bm{s}_0)$ is the initial state distribution, and $\gamma\in[0,1)$ is a discount factor. In the sequel, we denote the intrinsic reward by $\hat{r}$ to distinguish it from the extrinsic reward. Denote by $\pi(\bm{a}|\bm{s})$ the policy of the agent, the objective of reinforcement learning is to learn a policy that maximizes the expected discounted return $R_{t}=\mathbb{E}_{\pi}\left[\sum_{k=0}^{T}\gamma^{k}\check{r}_{t+k+1}\right]$.

%
\subsection{\renyi Divergence}
In this paper, we introduce \renyi divergence to evaluate the visitation discrepancy between different episodes, which is a generalized and flexible distance measure. In particular, \renyi divergence of order $1$ equals Kullback–Leibler divergence and \renyi divergence of order $0.5$ is a function of the square Hellinger distance \citep{van2014renyi}. Let $X\in\mathbb{R}^{d}$ be a random vector that has two density functions $p(\bm{x}),q(\bm{x})$ with respect to Lebesgue measure on $\mathbb{R}^{d}$, and let $\{\bm{x}\in\mathbb{R}^{d}:p(\bm{x}),q(\bm{x})>0\}$ be the support set. The \renyi divergence of order $\alpha\in\mathbb{R}\setminus\{1\}$ is defined as
\begin{equation}
	D_{\alpha}(p \Vert q)=\frac{1}{\alpha-1}\log\int_{\mathcal{X}}p^{\alpha}(\bm{x})q^{1-\alpha}(\bm{x})d\bm{x}.
\end{equation}
Since it is non-trivial to evaluate the metric when $p$ and $q$ are unavailable, the following $k$-NN estimator can be employed for efficient estimation. Let $\mathcal{X}=\{X_i\}_{i=1}^{N}$ be i.i.d samples from the distribution with density $p$ and $\mathcal{Y}=\{Y_i\}_{i=1}^{M}$ be i.i.d samples from the distribution with density $q$, we estimate the \renyi divergence as follows \citep{poczos2011estimation, kandasamy2015nonparametric}:
\begin{equation}\label{eq: estimator}
	\hat{D}^{k, N, M}_{\alpha}(p \Vert q)=\frac{1}{\alpha-1}\log\biggl\{\frac{1}{N}\sum_{i=1}^{N}\left[\frac{(N-1)\mu_{k}(X_i,\mathcal{X})}{M\nu_{k}(X_i,\mathcal{Y})}\right]^{1-\alpha}C_{k,\alpha}\biggr\},
\end{equation}
where $\mu_{k}(X_i,\mathcal{X})$ is the Euclidean distance between $X_i$ and its $k$-th nearest neighbor in $\mathcal{X}$, $\nu_{k}(X_i,\mathcal{Y})$ is the Euclidean distance between $X_i$ and its $k$-th nearest neighbor in $\mathcal{Y}$, $C_{k,\alpha}=\frac{\Gamma(k)^{2}}{\Gamma(k-\alpha+1)\Gamma(k+\alpha-1)}$, $k>|\alpha-1|$, and $\Gamma(\cdot)$ is the Gamma function. 

\section{REVD}
In this section, we present the REVD: \textbf{R}ewarding \textbf{E}pisodic \textbf{V}isitation \textbf{D}iscrepancy, a computation-efficient framework for providing powerful and sustainable exploration incentives in reinforcement learning. The key insight of \ourmethod is to encourage exploration by providing intrinsic rewards based on the episodic visitation discrepancy. Consider the on-policy learning scenario in which the policy is updated at the end of each episode, and let $\tilde{\pi}$ (the former one) and $\pi$ denote the learned policies of two consecutive episodes. The episodic visitation discrepancy can be evaluated using the \renyi divergence:
\begin{equation}\label{eq:div sd}
	D_{\alpha}(\rho^{\pi}(\bm{s})\Vert\rho^{\tilde{\pi}}(\bm{s})),
\end{equation}
where $\rho^{\pi}$ is the state distribution induced by the policy $\pi$. Therefore, if the agent finds novel states as compared to the previous episode, it will lead to a large divergence. This metric reflects the change of episode-level exploration status. In the exploration phase of reinforcement learning, the agent is expected to visit as many distinct states as possible within limited iterations. It is feasible to encourage exploration by maximizing $D_{\alpha}(\rho^{\pi}\Vert\rho^{\tilde{\pi}})$.

\subsection{Intrinsic Reward Design}
Next, we transform Eq.~(\ref{eq:div sd}) into computable intrinsic rewards. Denote by $\{\tilde{\bm{s}}_{i}\}_{i=1}^{T}$ and $\{\bm{s}_{i}\}_{i=1}^{T}$ the visited states sets of two consecutive episodes, the Eq.~(\ref{eq:div sd}) can be approximated using Eq.~(\ref{eq: estimator}) as
\begin{equation}
\begin{aligned}
		\hat{D}_{\alpha}^{k, T}(\rho^{\pi}\Vert\rho^{\tilde{\pi}})&=\frac{1}{\alpha-1}\log\biggr\{\frac{1}{T}\sum_{i=1}^{T}\left[\frac{(T-1)\mu_{k}(\bm{e}_{i},\mathcal{E})}{T\nu_{k}(\bm{e}_{i}, \tilde{\mathcal{E}})}\right]^{1-\alpha}C_{k,\alpha}\biggr\}\\
	&\propto \frac{1}{T}\sum_{i=1}^{T}\left[\frac{\nu_{k}(\bm{e}_{i}, \tilde{\mathcal{E}})}{\mu_{k}(\bm{e}_{i},\mathcal{E})}\right]^{1-\alpha},
\end{aligned}
\end{equation}
where $\bm{e}_{i}=f_{\bm \phi}(\bm{s}_i)$ is an encoding vector produced by a fixed and randomly-initialized encoder, $\mathcal{E}=\{f_{\bm \phi}(\bm{s}_i)\}_{i=1}^{T}$, and $\alpha\in(0,1)$. 
Therefore, we define the intrinsic reward that takes each transition as a particle:
\begin{equation}
	\hat{r}(\bm{s}_{i})=\left[\nu_{k}(\bm{e}_{i}, \tilde{\mathcal{E}})/(\mu_{k}(\bm{e}_{i}, \mathcal{E})+\epsilon)\right]^{1-\alpha},
\end{equation}
where $\epsilon$ is a fixed and small constant. A critical problem of this formulation is that the agent may obtain more intrinsic rewards by lingering in a small area. To address the problem, we consider reformulating the intrinsic reward as 
\begin{equation}\label{eq:intrinsic reward}
	\hat{r}(\bm{s}_{i})=L(\mathcal{E})\cdot\left[\nu_{k}(\bm{e}_{i}, \tilde{\mathcal{E}})/(\mu_{k}(\bm{e}_{i}, \mathcal{E})+\epsilon)\right]^{1-\alpha},
\end{equation}
where $L( \mathcal{E})=\tanh\left(\frac{1}{T}\sum_{i=0}^{T}\mu_{1}(\bm{e}_{i},\mathcal{E})\right)$ is a scaling coefficient. 

\begin{algorithm}[t]
	\caption{REVD}
	\label{algo:revd}
	\begin{algorithmic}[1]
		\STATE Initialize the encoder $f_{\bm\phi}$ and policy network $\pi_{\bm{\theta}}$;
		\STATE Initialize the maximum episodes $E$, order $\alpha$, coefficient $\lambda_{0}$, decay rate $\kappa$, constant $\epsilon$, and replay buffer $\tilde{\mathcal{E}},\mathcal{E}\leftarrow\emptyset$;
		\FOR{$\ell=1,\dots,E$}
		\STATE Collect the trajectory $\bm{s}_{0},\bm{a}_{0},\dots,\bm{a}_{T-1},\bm{s}_{T}$;
		\STATE Compute the encoding vectors of $\{\bm{s}_t\}_{t=0}^{T}$ and $\mathcal{E}\leftarrow\mathcal{E}\cup\{\bm{e}_t\}_{t=0}^{T}$;
		\IF{$\ell=1$}
		\STATE Set the intrinsic reward as $0$;
		\STATE $r^{\rm total}_{t}\leftarrow\check{r}(\bm{s}_{t},\bm{a}_{t})$;
		\ELSE
		\STATE Compute the scaling coefficient $L(\mathcal{E})$;
		\STATE Compute $\hat{r}(\bm{s}_{t})\leftarrow L(\mathcal{E})\cdot\left[\nu_{k}(\bm{e}_{t}, \tilde{\mathcal{E}})/\mu_{k}(\bm{e}_{t}, \mathcal{E})+\epsilon\right]^{1-\alpha}$;
		\STATE $r^{\rm total}_{t}\leftarrow\check{r}(\bm{s}_{t},\bm{a}_{t})+\lambda_{\ell}\cdot\hat{r}(\bm{s}_{t})$;
		\ENDIF
		\STATE Update $\lambda_{\ell}=\lambda_{0}(1-\kappa)^{\ell}$;
		\STATE Update the policy network with transitions $\{\bm{s}_{t},\bm{a}_{t},\bm{s}_{t+1},r_{t}^{\rm total}\}$ using any on-policy reinforcement learning algorithm;
		\STATE $\tilde{\mathcal{E}}\leftarrow\mathcal{E},\mathcal{E}\leftarrow\emptyset$.
		\ENDFOR
	\end{algorithmic}
\end{algorithm}

\begin{figure}[h]
	\centering
	\includegraphics[width=0.9\linewidth]{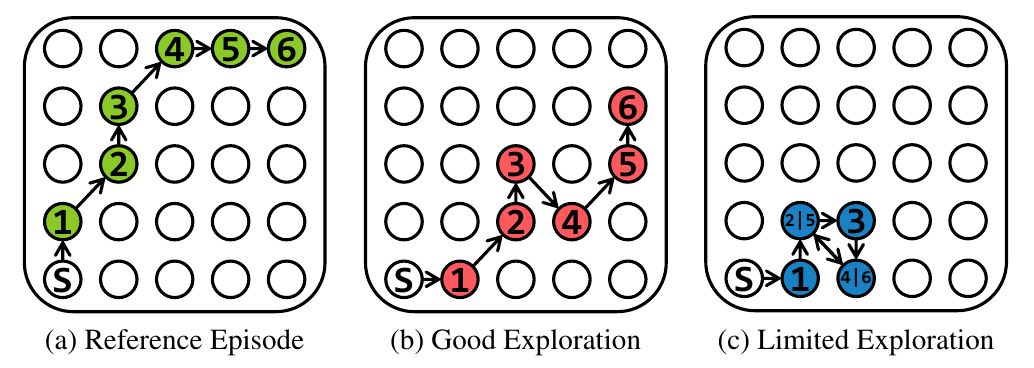}
	\caption{Exploration comparison among three episodes. It is natural to find that Episode (b) achieves better exploration because it visited more distinct states when compared to the reference episode.}
	\label{fig:exploration comparison}
\end{figure}

Here we use the average distance between the states and their nearest neighbors to characterize the intra-episode difference. As a result, the agent lingering in a small region leads to a small $L$, and the derived intrinsic reward will be decreased. We use the following example to further demonstrate the effectiveness of $L$. Figure \ref{fig:exploration comparison} illustrates the trajectories of three episodes. Consider $k=3, \epsilon=0.0001$ and $\alpha=0.5$, the average intrinsic reward computed by Eq.~(\ref{eq:intrinsic reward}) of episode (b) and episode (c) is 0.89 and 0.67, respectively. In episode (b), the agent visited unseen and distinct states when compared to the reference episode, which emphasizes both the intra-episode exploration and exploration across episodes. In contrast, the agent visited novel and repetitive states in episode (c), which will be punished because of low intra-episode difference. 

\subsection{Training Objective}
Equipped with the intrinsic reward, the total reward of each transition $(\bm{s}_{t},\bm{a}_{t},\bm{s}_{t+1})$ is computed as
\begin{equation}\label{eq:total reward}
	r_{t}^{\rm total}=\check{r}(\bm{s}_{t},\bm{a}_{t})+\lambda_{t}\cdot\hat{r}(\bm{s}_{t}),
\end{equation}
where $\lambda_{t}=\lambda_{0}(1-\kappa)^{t}\geq 0$ is a weighting coefficient that determines the exploration preference, where $\kappa$ is a decay rate. In particular, this intrinsic reward can be used to perform unsupervised pre-training in the absence of extrinsic rewards $\check{r}$. Then the pre-trained policy can be employed in the downstream task adaptation with extrinsic rewards. Letting $\pi_{\bm{\theta}}$ denote the policy represented by a neural network with parameters $\bm{\theta}$, the training objective is to maximize the expected discounted return $\mathbb{E}_{\pi_{\bm{\theta}}}\left[\sum_{k=0}^{T}\gamma^{k}r_{t+k+1}^{\rm total}\right]$. Finally, the detailed workflow of REVD is summarized in Algorithm~\ref{algo:revd}.

\section{Experiments}
We designed the experiments to answer the following questions:
\begin{itemize}
	\item Can REVD improve the sample efficiency and policy performance of reinforcement learning algorithms in complex environments with high-dimensional observations? (See Figure \ref{fig:atari eps return 1})
	\item How does REVD compare to computation-efficient exploration approaches and other exploration approaches empowered by auxiliary models? (See Figure~\ref{fig:atari eps return 2} and Figure~\ref{fig:bullet eps return 2})
	\item Can REVD achieve remarkable performance in sparse-reward setting? (See Figure~\ref{fig:atari eps return 3} and Figure~\ref{fig:bullet eps return 3})
	\item Can REVD also improve the sample efficiency in continuous control tasks? (See Figure~\ref{fig:bullet eps return 1}, Figure~\ref{fig:bullet eps return 2}, and Figure~\ref{fig:bullet eps return 3})
\end{itemize}

\subsection{Atari Games}

\subsubsection{Setup}
We first tested REVD on six Atari games with graphic observations and discrete action space \citep{brockman2016openai}. To evaluate the sample efficiency of REVD, two representative reinforcement learning algorithms were introduced to serve as baselines, namely the Advantage Actor-Critic (A2C) and Proximal Policy Optimization (PPO) \citep{schulman2017proximal, mnih2016asynchronous}. For comparison with computation-efficient exploration, we selected RE3 \citep{seo2021state} as the benchmarking scheme, which maximizes the state entropy using a non-parametric entropy estimator. For comparison with exploration approaches empowered by auxiliary models, we selected RIDE \citep{raileanu2020ride} that uses the difference between consecutive states as intrinsic rewards. A brief introduction of RE3 and RIDE can be found in Appendix \ref{appendix:bmm}. The following results were obtained by setting $k=5, \alpha=0.5, \lambda_{0}=0.1, \kappa=0.00001$ and $\epsilon=0.0001$, and more details are provided in Appendix \ref{appendix:dage}.

\subsubsection{Results}
\begin{figure}[h]
	\centering
	\includegraphics[width=1.0\linewidth]{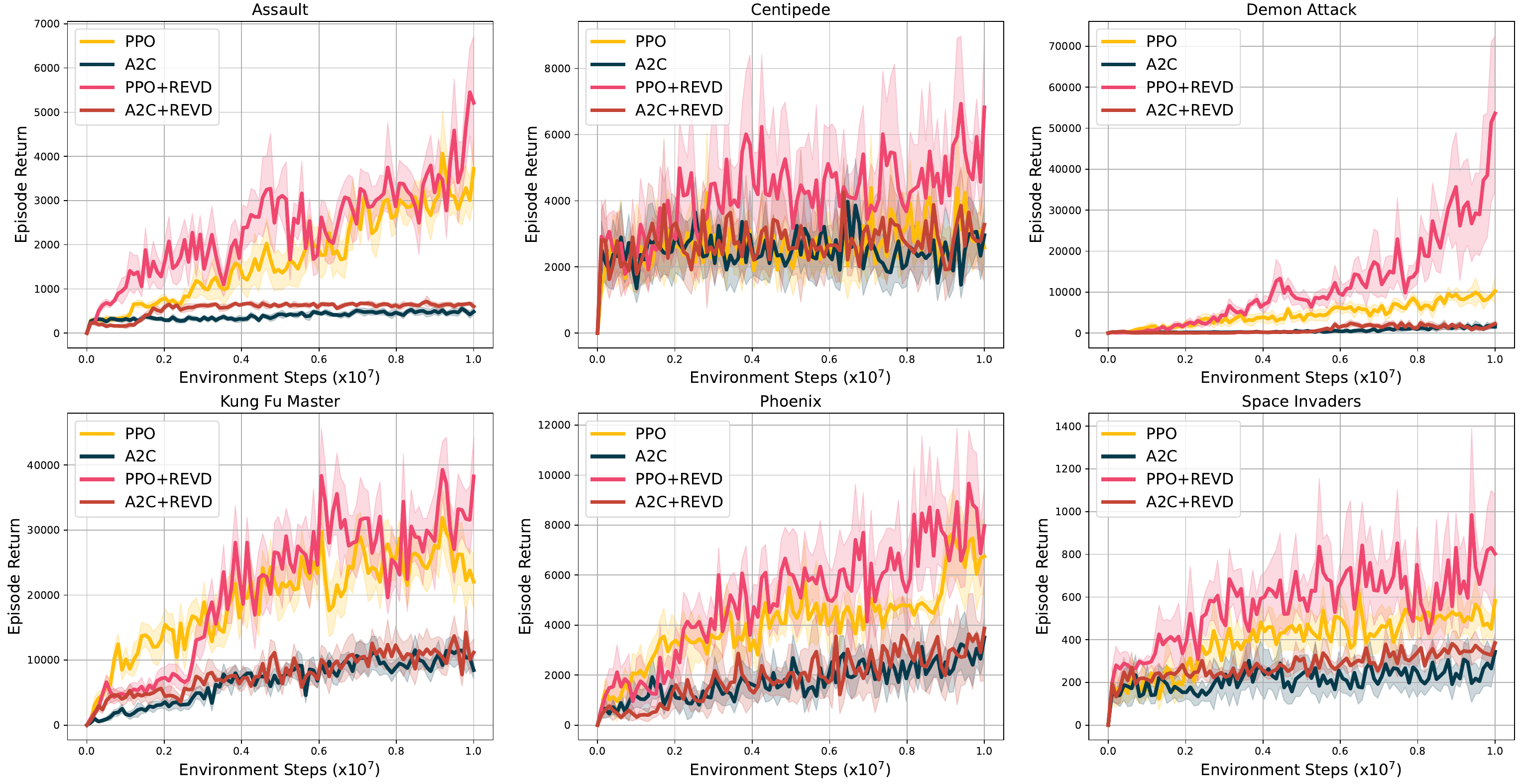}
	\caption{Performance on six Atari games over ten random seeds. REVD significantly promotes the sample efficiency with simple architecture and higher efficiency. The solid line and shaded regions represent the mean and standard deviation, respectively.}
	\label{fig:atari eps return 1}
\end{figure}
Figure \ref{fig:atari eps return 1} illustrates the comparison of the average episode return of six Atari games. It is obvious that REVD significantly improved the sample efficiency of PPO in various tasks. In particular, PPO+REVD achieved an average episode return of 52,000 in \textit{Demon Attack} task, producing a giant performance gain when compared to the vanilla PPO agent. In contrast, the vanilla A2C agent achieved low performance in these tasks. But REVD also promoted its sample efficiency in most tasks, demonstrating the effectiveness of REVD.

\begin{figure}[h]
	\centering
	\includegraphics[width=1.0\linewidth]{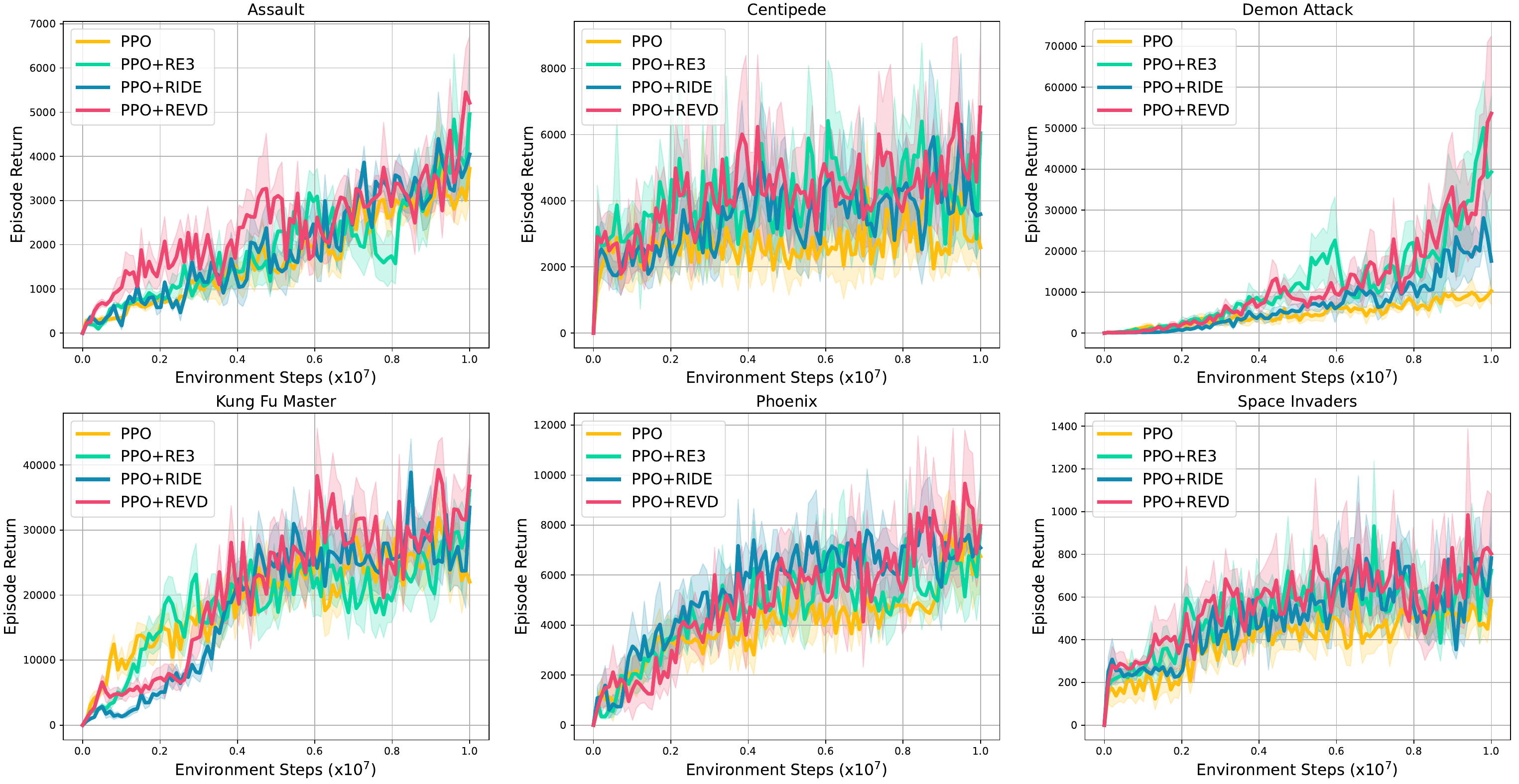}
	\caption{Performance on six Atari games over ten random seeds. PPO+REVD outperforms the benchmarking schemes in all the tasks. The solid line and shaded regions represent the mean and standard deviation, respectively.}
	\label{fig:atari eps return 2}
\end{figure}

Next, we compared REVD with other exploration methods and considered combining them with PPO agent, \ie, PPO+RE3 and PPO+RIDE. As shown in Figure \ref{fig:atari eps return 2}, PPO+REVD also achieved the highest policy performance in all tasks. As a computation-efficient method, PPO+RE3 achieved the second highest performance in four tasks. In contrast, PPO+RIDE achieved the second highest performance in two tasks. It is clear that the exploration-empowered agent consistently obtained higher performance than the vanilla PPO agent. Figure \ref{fig:atari eps return 1} and Figure \ref{fig:atari eps return 2} demonstrate that REVD can effectively promote the sample efficiency and policy performance of reinforcement learning algorithms without introducing any auxiliary models and representation learning.

\begin{figure}
\includegraphics[width=1.0\linewidth]{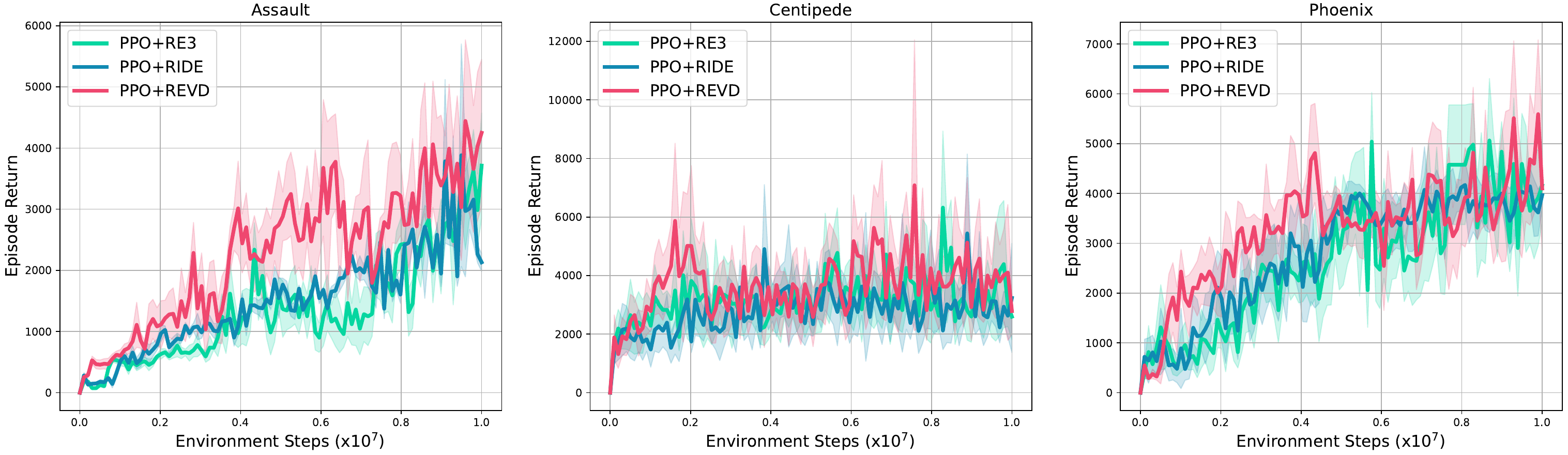}
	\caption{Performance on six Atari games with sparse rewards over ten random seeds. The solid line and shaded area represent the mean and standard deviation, respectively.}
	\label{fig:atari eps return 3}
\end{figure}

Furthermore, we tested REVD considering the sparse-reward setting, in which the original extrinsic rewards will be randomly set as zero. Figure \ref{fig:atari eps return 3} demonstrates that REVD can still achieve superior sample efficiency in sparse-reward setting. In particular, REVD produced an oscillatory learning curve in \textit{Assault} task, which indicates that REVD provided aggressive exploration incentives. The combination of PPO and RE3 also obtained remarkable performance in all the games, which demonstrates the of high effectiveness and efficiency of computation-efficient exploration.

\subsection{PyBullet Robotics Environments}
\begin{figure}[h]
	\centering
	\includegraphics[width=0.95\linewidth]{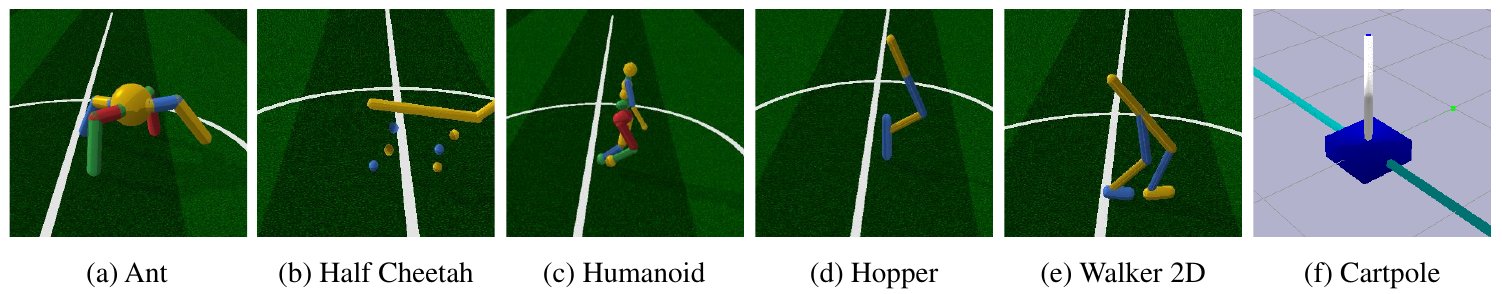}
	\caption{Render images of locomotion tasks from PyBullet Robotics Environments. For instance, the goal of \textit{Humanoid} task is to walk forward as fast as possible without falling over.}
	\label{fig:bullet games}
\end{figure}

\subsubsection{Setup}
Next, we tested REVD using physical simulation tasks of PyBullet Robotics Environments (See Figure~\ref{fig:bullet games}) \cite{coumans2016pybullet}. We also selected A2C and PPO as the baselines and compared the sample efficiency of REVD with RE3 and RIDE. The following results were obtained by setting $k=3, \alpha=0.5, \lambda_{0}=0.1, \kappa=0.00001$ and $\epsilon=0.0001$, and more details are provided in Appendix \ref{appendix:ddcse}.

\subsection{Results}
\begin{figure}[h]
	\centering
\includegraphics[width=1.0\linewidth]{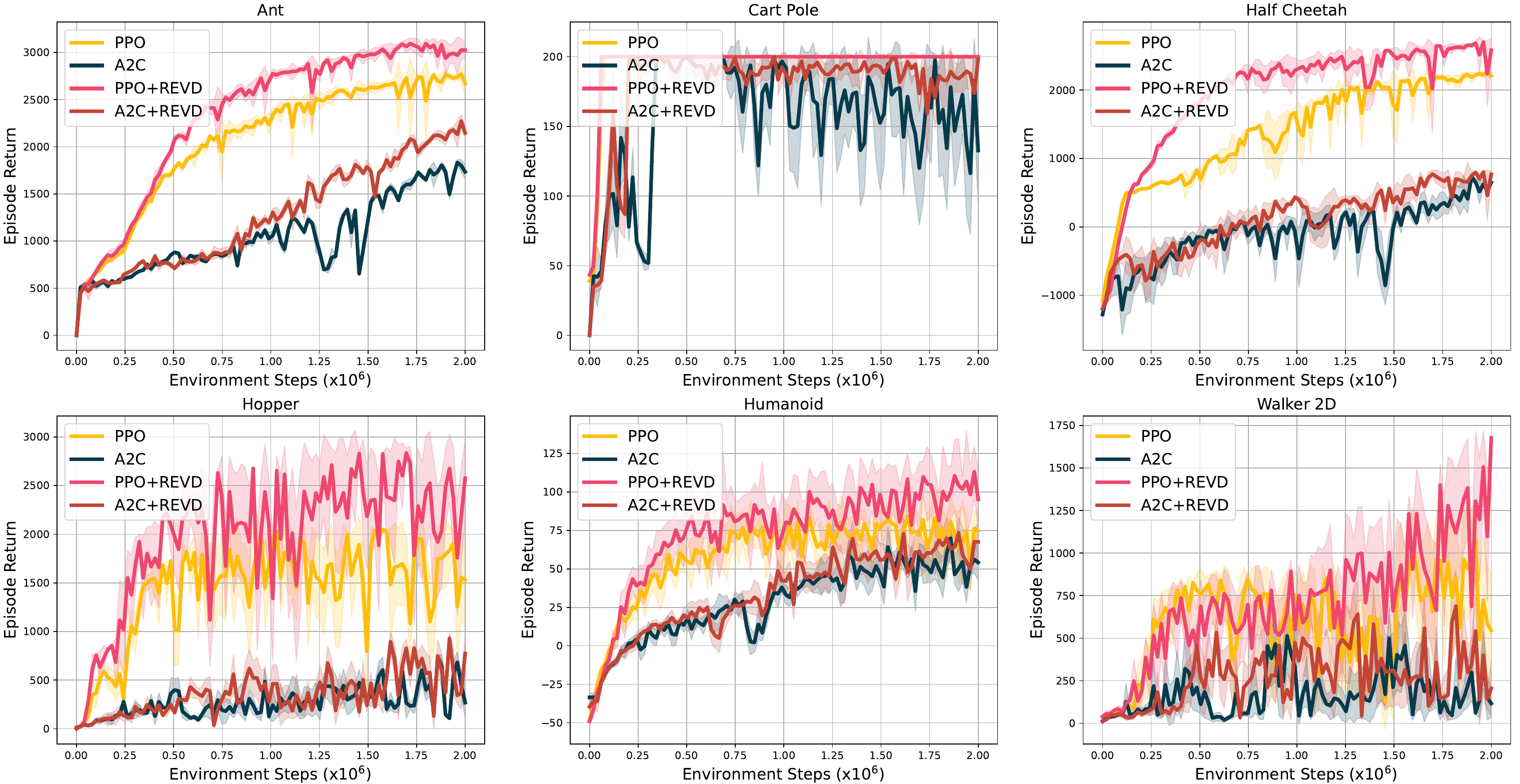}
	\caption{Performance on six PyBullet tasks over ten random seeds. REVD significantly improves the sample efficiency of PPO and A2C. The solid line and shaded regions represent the mean and standard deviation, respectively.}
	\label{fig:bullet eps return 1}
\end{figure}

Figure \ref{fig:bullet eps return 1} illustrates the comparison of the average episode return of six PyBullet tasks. REVD also significantly promoted the sample efficiency of PPO and A2C in various tasks. In \textit{Walker 2D} task, PPO+REVD obtained the highest average episode return, while the vanilla PPO agent prematurely fell into local optima. In \textit{Cart Pole} task, PPO+REVD solved the problem using the minimum environment steps. For \textit{Hopper} and \textit{Humanoid} task, PPO+REVD achieved the highest convergence rate, demonstrating the powerful exploration capability of REVD. In contrast, A2C achieved low performance in \textit{Half Cheetah} and \textit{Humanoid} task. But A2C+REVD also achieved higher performance than the vanilla A2C agent. 

\begin{figure}[h]
	\centering
	\includegraphics[width=1.0\linewidth]{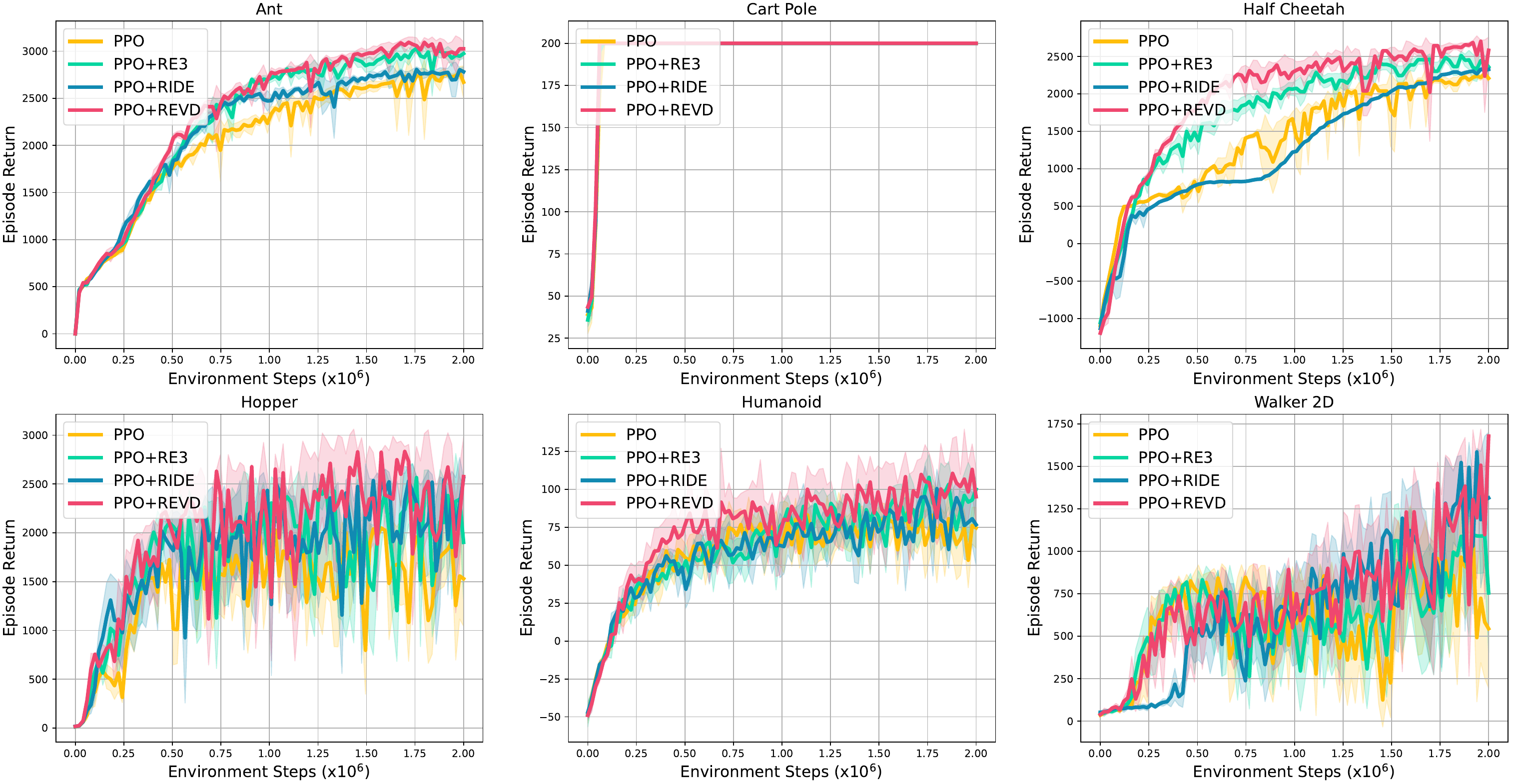}
	\caption{Performance on six PyBullet tasks over ten random seeds. PPO+REVD also outperforms the benchmarking schemes in all the tasks. The solid line and shaded regions represent the mean and standard deviation, respectively.}
	\label{fig:bullet eps return 2}
\end{figure}

\begin{figure}[h]
	\centering
	\includegraphics[width=1.0\linewidth]{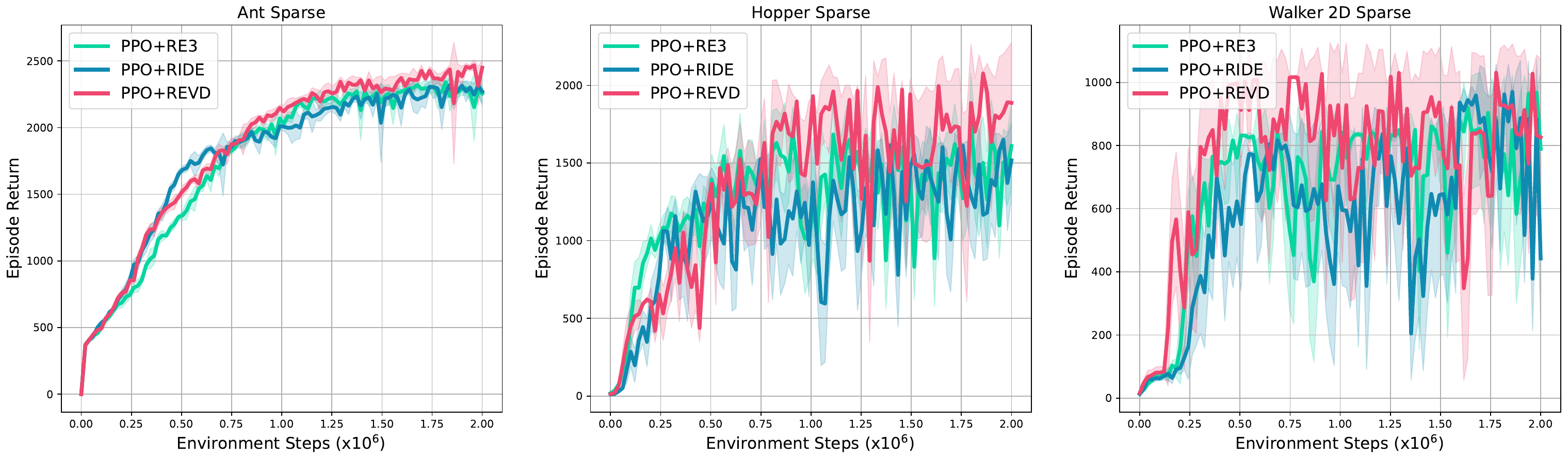}
	\caption{Performance on six PyBullet tasks with sparse rewards over ten random seeds. The solid line and shaded area represent the mean and standard deviation, respectively.}
	\label{fig:bullet eps return 3}
\end{figure}

Figure \ref{fig:bullet eps return 2} illustrates the comparison of REVD with other exploration methods, and PPO+REVD consistently achieved the highest sample efficiency in most tasks. Finally, we also tested REVD in the sparse-reward setting. Figure \ref{fig:bullet eps return 3} demonstrates that REVD also outperformed the benchmarking methods with higher policy performance. Therefore, REVD can effectively improve the sample efficiency in both the continuous and discrete control tasks.

%

\section{Conclusion}
In this paper, we propose a novel and computation-efficient exploration method entitled REVD, which provides high-quality intrinsic rewards without auxiliary models or representation learning. Since REVD computes intrinsic rewards based on episodic visitation discrepancy, it can overcome the problem of vanishing rewards and provides sustainable exploration incentives. In particular, REVD provides a quantified metric that reflects the episode-level exploration change. Finally, the REVD is tested on Atari games and PyBullet Robotics Environments. The simulation results demonstrate that REVD and can significantly improve the sample efficiency and outperform the benchmarking schemes. It can be envisaged that the computation-efficient exploration will be widely applied in arbitrary tasks due to its high efficiency and convenience. This work is expected to facilitate and inspire more consequent research on computation-efficient exploration methods.



\begin{thebibliography}{38}
\providecommand{\natexlab}[1]{#1}
\providecommand{\url}[1]{\texttt{#1}}
\expandafter\ifx\csname urlstyle\endcsname\relax
  \providecommand{\doi}[1]{doi: #1}\else
  \providecommand{\doi}{doi: \begingroup \urlstyle{rm}\Url}\fi

\bibitem[Agarap(2018)]{agarap2018deep}
Abien~Fred Agarap.
\newblock Deep learning using rectified linear units (relu).
\newblock \emph{arXiv preprint arXiv:1803.08375}, 2018.

\bibitem[Badia et~al.(2020)Badia, Sprechmann, Vitvitskyi, Guo, Piot,
  Kapturowski, Tieleman, Arjovsky, Pritzel, Bolt, and Blundell]{badia2020never}
Adrià~Puigdomènech Badia, Pablo Sprechmann, Alex Vitvitskyi, Daniel Guo,
  Bilal Piot, Steven Kapturowski, Olivier Tieleman, Martin Arjovsky, Alexander
  Pritzel, Andrew Bolt, and Charles Blundell.
\newblock Never give up: Learning directed exploration strategies.
\newblock In \emph{Proceedings of the International Conference on Learning
  Representations}, 2020.

\bibitem[Bellemare et~al.(2016)Bellemare, Srinivasan, Ostrovski, Schaul,
  Saxton, and Munos]{bellemare2016unifying}
Marc Bellemare, Sriram Srinivasan, Georg Ostrovski, Tom Schaul, David Saxton,
  and Remi Munos.
\newblock Unifying count-based exploration and intrinsic motivation.
\newblock \emph{Proceedings of Advances in Neural Information Processing
  Systems}, 29:\penalty0 1471--1479, 2016.

\bibitem[Bellman(1957)]{bellman1957markovian}
Richard Bellman.
\newblock A markovian decision process.
\newblock \emph{Journal of mathematics and mechanics}, pp.\  679--684, 1957.

\bibitem[Brockman et~al.(2016)Brockman, Cheung, Pettersson, Schneider,
  Schulman, Tang, and Zaremba]{brockman2016openai}
Greg Brockman, Vicki Cheung, Ludwig Pettersson, Jonas Schneider, John Schulman,
  Jie Tang, and Wojciech Zaremba.
\newblock Open{AI} gym.
\newblock \emph{arXiv preprint arXiv:1606.01540}, 2016.

\bibitem[Burda et~al.(2019{\natexlab{a}})Burda, Edwards, Pathak, Storkey,
  Darrell, and Efros]{burda2018large}
Yuri Burda, Harri Edwards, Deepak Pathak, Amos Storkey, Trevor Darrell, and
  Alexei~A Efros.
\newblock Large-scale study of curiosity-driven learning.
\newblock \emph{Proceedings of the International Conference on Learning
  Representations}, pp.\  1--17, 2019{\natexlab{a}}.

\bibitem[Burda et~al.(2019{\natexlab{b}})Burda, Edwards, Storkey, and
  Klimov]{burda2018exploration}
Yuri Burda, Harrison Edwards, Amos Storkey, and Oleg Klimov.
\newblock Exploration by random network distillation.
\newblock \emph{Proceedings of the 7th International Conference on Learning
  Representations}, pp.\  1--17, 2019{\natexlab{b}}.

\bibitem[Coumans \& Bai(2016–2018)Coumans and Bai]{coumans2016pybullet}
Erwin Coumans and Yunfei Bai.
\newblock Pybullet, a python module for physics simulation for games, robotics
  and machine learning.
\newblock \emph{URL http://pybullet.org}, 2016–2018.

\bibitem[Houthooft et~al.(2016)Houthooft, Chen, Duan, Schulman, De~Turck, and
  Abbeel]{houthooft2016vime}
Rein Houthooft, Xi~Chen, Yan Duan, John Schulman, Filip De~Turck, and Pieter
  Abbeel.
\newblock Vime: Variational information maximizing exploration.
\newblock \emph{Proceedings of Advances in neural information processing
  systems}, 29:\penalty0 1109--1117, 2016.

\bibitem[Ioffe \& Szegedy(2015)Ioffe and Szegedy]{ioffe2015batch}
Sergey Ioffe and Christian Szegedy.
\newblock Batch normalization: Accelerating deep network training by reducing
  internal covariate shift.
\newblock In \emph{Proceedings of the International conference on machine
  learning}, pp.\  448--456, 2015.

\bibitem[Kaelbling et~al.(1998)Kaelbling, Littman, and
  Cassandra]{kaelbling1998planning}
Leslie~Pack Kaelbling, Michael~L Littman, and Anthony~R Cassandra.
\newblock Planning and acting in partially observable stochastic domains.
\newblock \emph{Artificial intelligence}, 101\penalty0 (1-2):\penalty0 99--134,
  1998.

\bibitem[Kandasamy et~al.(2015)Kandasamy, Krishnamurthy, Poczos, Wasserman,
  et~al.]{kandasamy2015nonparametric}
Kirthevasan Kandasamy, Akshay Krishnamurthy, Barnabas Poczos, Larry Wasserman,
  et~al.
\newblock Nonparametric von mises estimators for entropies, divergences and
  mutual informations.
\newblock \emph{Advances in Neural Information Processing Systems}, 28, 2015.

\bibitem[Kingma \& Welling(2014)Kingma and Welling]{kingma2013auto}
Diederik~P Kingma and Max Welling.
\newblock Auto-encoding variational bayes.
\newblock \emph{Proceedings of the International Conference on Learning
  Representations}, 2014.

\bibitem[LeCun et~al.(2015)LeCun, Bengio, and Hinton]{lecun2015deep}
Yann LeCun, Yoshua Bengio, and Geoffrey Hinton.
\newblock Deep learning.
\newblock \emph{nature}, 521\penalty0 (7553):\penalty0 436--444, 2015.

\bibitem[Machado et~al.(2020)Machado, Bellemare, and Bowling]{machado2020count}
Marlos~C Machado, Marc~G Bellemare, and Michael Bowling.
\newblock Count-based exploration with the successor representation.
\newblock In \emph{Proceedings of the AAAI Conference on Artificial
  Intelligence}, volume~34, pp.\  5125--5133, 2020.

\bibitem[Martin et~al.(2017)Martin, S., Everitt, and Hutter]{ijcai2017p0344}
Jarryd Martin, Suraj~Narayanan S., Tom Everitt, and Marcus Hutter.
\newblock Count-based exploration in feature space for reinforcement learning.
\newblock In \emph{Proceedings of the Twenty-Sixth International Joint
  Conference on Artificial Intelligence, {IJCAI-17}}, pp.\  2471--2478, 2017.
\newblock \doi{10.24963/ijcai.2017/344}.
\newblock URL \url{https://doi.org/10.24963/ijcai.2017/344}.

\bibitem[Mnih et~al.(2015)Mnih, Kavukcuoglu, Silver, Rusu, Veness, Bellemare,
  Graves, Riedmiller, Fidjeland, Ostrovski, et~al.]{mnih2015human}
Volodymyr Mnih, Koray Kavukcuoglu, David Silver, Andrei~A Rusu, Joel Veness,
  Marc~G Bellemare, Alex Graves, Martin Riedmiller, Andreas~K Fidjeland, Georg
  Ostrovski, et~al.
\newblock Human-level control through deep reinforcement learning.
\newblock \emph{nature}, 518\penalty0 (7540):\penalty0 529--533, 2015.

\bibitem[Mnih et~al.(2016)Mnih, Badia, Mirza, Graves, Lillicrap, Harley,
  Silver, and Kavukcuoglu]{mnih2016asynchronous}
Volodymyr Mnih, Adria~Puigdomenech Badia, Mehdi Mirza, Alex Graves, Timothy
  Lillicrap, Tim Harley, David Silver, and Koray Kavukcuoglu.
\newblock Asynchronous methods for deep reinforcement learning.
\newblock In \emph{International conference on machine learning}, pp.\
  1928--1937. PMLR, 2016.

\bibitem[Mutti et~al.(2021)Mutti, Pratissoli, and Restelli]{mutti2021task}
Mirco Mutti, Lorenzo Pratissoli, and Marcello Restelli.
\newblock Task-agnostic exploration via policy gradient of a non-parametric
  state entropy estimate.
\newblock In \emph{Proceedings of the AAAI Conference on Artificial
  Intelligence}, volume~35, pp.\  9028--9036, 2021.

\bibitem[Oh et~al.(2015)Oh, Guo, Lee, Lewis, and Singh]{oh2015action}
Junhyuk Oh, Xiaoxiao Guo, Honglak Lee, Richard~L Lewis, and Satinder Singh.
\newblock Action-conditional video prediction using deep networks in atari
  games.
\newblock \emph{Advances in neural information processing systems}, 28, 2015.

\bibitem[Ostrovski et~al.(2017)Ostrovski, Bellemare, Oord, and
  Munos]{ostrovski2017count}
Georg Ostrovski, Marc~G Bellemare, A{\"a}ron Oord, and R{\'e}mi Munos.
\newblock Count-based exploration with neural density models.
\newblock In \emph{Proceedings of the International Conference on Machine
  Learning}, pp.\  2721--2730, 2017.

\bibitem[Oudeyer \& Kaplan(2008)Oudeyer and Kaplan]{oudeyer2008can}
Pierre-Yves Oudeyer and Frederic Kaplan.
\newblock How can we define intrinsic motivation?
\newblock In \emph{the 8th International Conference on Epigenetic Robotics:
  Modeling Cognitive Development in Robotic Systems}. Lund University Cognitive
  Studies, Lund: LUCS, Brighton, 2008.

\bibitem[Oudeyer et~al.(2007)Oudeyer, Kaplan, and Hafner]{oudeyer2007intrinsic}
Pierre-Yves Oudeyer, Frdric Kaplan, and Verena~V Hafner.
\newblock Intrinsic motivation systems for autonomous mental development.
\newblock \emph{IEEE transactions on evolutionary computation}, 11\penalty0
  (2):\penalty0 265--286, 2007.

\bibitem[Pathak et~al.(2017)Pathak, Agrawal, Efros, and
  Darrell]{pathak2017curiosity}
Deepak Pathak, Pulkit Agrawal, Alexei~A Efros, and Trevor Darrell.
\newblock Curiosity-driven exploration by self-supervised prediction.
\newblock In \emph{Proceedings of the IEEE Conference on Computer Vision and
  Pattern Recognition Workshops}, pp.\  16--17, 2017.

\bibitem[P{\'o}czos \& Schneider(2011)P{\'o}czos and
  Schneider]{poczos2011estimation}
Barnab{\'a}s P{\'o}czos and Jeff Schneider.
\newblock On the estimation of alpha-divergences.
\newblock In \emph{Proceedings of the Fourteenth International Conference on
  Artificial Intelligence and Statistics}, pp.\  609--617. JMLR Workshop and
  Conference Proceedings, 2011.

\bibitem[Raileanu \& Rocktäschel(2020)Raileanu and
  Rocktäschel]{raileanu2020ride}
Roberta Raileanu and Tim Rocktäschel.
\newblock Ride: Rewarding impact-driven exploration for procedurally-generated
  environments.
\newblock In \emph{Proceedings of the International Conference on Learning
  Representations}, 2020.
\newblock URL \url{https://openreview.net/forum?id=rkg-TJBFPB}.

\bibitem[Schmidhuber(1991)]{schmidhuber1991possibility}
J{\"u}rgen Schmidhuber.
\newblock A possibility for implementing curiosity and boredom in
  model-building neural controllers.
\newblock In \emph{Proc. of the international conference on simulation of
  adaptive behavior: From animals to animats}, pp.\  222--227, 1991.

\bibitem[Schulman et~al.(2015)Schulman, Moritz, Levine, Jordan, and
  Abbeel]{schulman2015high}
John Schulman, Philipp Moritz, Sergey Levine, Michael Jordan, and Pieter
  Abbeel.
\newblock High-dimensional continuous control using generalized advantage
  estimation.
\newblock \emph{Proceedings of the International Conference on Learning
  Representations}, 2015.

\bibitem[Schulman et~al.(2017)Schulman, Wolski, Dhariwal, Radford, and
  Klimov]{schulman2017proximal}
John Schulman, Filip Wolski, Prafulla Dhariwal, Alec Radford, and Oleg Klimov.
\newblock Proximal policy optimization algorithms.
\newblock \emph{arXiv preprint arXiv:1707.06347}, 2017.

\bibitem[Seo et~al.(2021)Seo, Chen, Shin, Lee, Abbeel, and Lee]{seo2021state}
Younggyo Seo, Lili Chen, Jinwoo Shin, Honglak Lee, Pieter Abbeel, and Kimin
  Lee.
\newblock State entropy maximization with random encoders for efficient
  exploration.
\newblock In \emph{Proceedings of the 38th International Conference on Machine
  Learning}, pp.\  9443--9454, 2021.

\bibitem[Stadie et~al.(2015)Stadie, Levine, and
  Abbeel]{stadie2015incentivizing}
Bradly~C Stadie, Sergey Levine, and Pieter Abbeel.
\newblock Incentivizing exploration in reinforcement learning with deep
  predictive models.
\newblock \emph{arXiv preprint arXiv:1507.00814}, 2015.

\bibitem[Strehl \& Littman(2008)Strehl and Littman]{strehl2008analysis}
Alexander~L Strehl and Michael~L Littman.
\newblock An analysis of model-based interval estimation for markov decision
  processes.
\newblock \emph{Journal of Computer and System Sciences}, 74\penalty0
  (8):\penalty0 1309--1331, 2008.

\bibitem[Tang et~al.(2017)Tang, Houthooft, Foote, Stooke, Xi~Chen, Duan,
  Schulman, DeTurck, and Abbeel]{tang2017exploration}
Haoran Tang, Rein Houthooft, Davis Foote, Adam Stooke, OpenAI Xi~Chen, Yan
  Duan, John Schulman, Filip DeTurck, and Pieter Abbeel.
\newblock \# exploration: A study of count-based exploration for deep
  reinforcement learning.
\newblock \emph{Advances in neural information processing systems}, 30, 2017.

\bibitem[Van~Erven \& Harremos(2014)Van~Erven and Harremos]{van2014renyi}
Tim Van~Erven and Peter Harremos.
\newblock R{\'e}nyi divergence and kullback-leibler divergence.
\newblock \emph{IEEE Transactions on Information Theory}, 60\penalty0
  (7):\penalty0 3797--3820, 2014.

\bibitem[Yu et~al.(2020)Yu, Lyu, and Tsang]{yu2020intrinsic}
Xingrui Yu, Yueming Lyu, and Ivor Tsang.
\newblock Intrinsic reward driven imitation learning via generative model.
\newblock In \emph{Proceedings of the International Conference on Machine
  Learning}, pp.\  10925--10935, 2020.

\bibitem[Yuan et~al.(2022)Yuan, Pun, and Wang]{yuan2022renyi}
Mingqi Yuan, Man-On Pun, and Dong Wang.
\newblock R{\'e}nyi state entropy maximization for exploration acceleration in
  reinforcement learning.
\newblock \emph{IEEE Transactions on Artificial Intelligence}, 2022.

\bibitem[Zhang et~al.(2021{\natexlab{a}})Zhang, Cai, and
  Li]{zhang2021exploration}
Chuheng Zhang, Yuanying Cai, and Longbo Huang~Jian Li.
\newblock Exploration by maximizing {R}{\'e}nyi entropy for reward-free {RL}
  framework.
\newblock In \emph{Proceedings of the AAAI Conference on Artificial
  Intelligence}, pp.\  10859--10867, 2021{\natexlab{a}}.

\bibitem[Zhang et~al.(2021{\natexlab{b}})Zhang, Rashidinejad, Jiao, Tian,
  Gonzalez, and Russell]{zhang2021made}
Tianjun Zhang, Paria Rashidinejad, Jiantao Jiao, Yuandong Tian, Joseph~E
  Gonzalez, and Stuart Russell.
\newblock Made: Exploration via maximizing deviation from explored regions.
\newblock \emph{Advances in Neural Information Processing Systems},
  34:\penalty0 9663--9680, 2021{\natexlab{b}}.

\end{thebibliography}

\newpage
\appendix
\section{Benchmarking methods}\label{appendix:bmm}
\subsection{RE3} 
Given a trajectory $\{\bm{s}_{0},\bm{a}_{0},\dots,\bm{a}_{T-1},\bm{s}_{T}\}$ collected the agent, RE3 first uses a random encoder to encode the visited states before using a $k$-NN estimator to compute the state entropy \citep{seo2021state}. Denote by $\mathcal{E}=\{\bm{e}_{i}\}_{i=1}^{T}$ the encoding vectors, RE3 proposes to define the intrinsic reward as
\begin{equation}
\hat{r}(\bm{s}_{t})=\log\left(\mu_{k}(\bm{e}_{t},\mathcal{E})+1\right).
\end{equation}
Then the intrinsic reward can be used to online RL and unsupervised pre-training. 

\subsection{RIDE}
RIDE is built based on the ICM \citep{pathak2017curiosity}, which trains a forward and an inverse dynamics model to learn the representation of the state space. Denote by $\phi(\bm{s})$ the state representation, RIDE computes the intrinsic reward as the following Euclidean distance:
\begin{equation}
	\hat{r}(\bm{s}_t)=\frac{\Vert\phi(\bm{s}_{t+1})-\phi(\bm{s}_{t})\Vert_{2}}{\sqrt{N_{ep}(\bm{s}_{t+1})}},
\end{equation}
where $N_{ep}$ is the state visitation frequency during the current episode. $N_{ep}$ is used to discount the intrinsic reward, which prevents the agent from lingering in a sequence of states with a large difference in their embeddings. 

\section{Details on Atari Games Experiments}\label{appendix:dage}
\subsection{Environment Setting}
To handle the graphic observations, we stacked 4 consecutive frames as one state, and these frames were cropped to the size of (84, 84) to reduce the computational complexity. Moreover, we clip the extrinsic reward using a sign function to accelerate the training process. To simulate the sparse-reward scenario, we modified the original environment by adding a reward constraint, which will randomly set the value of extrinsic reward as zero.

\subsection{Experimental Setting for RL}
In this work, we used a PyTorch implementation of A2C and PPO that can be found in (\url{https://github.com/DLR-RM/stable-baselines3}). An identical policy network and value network were employed for all methods to make a fair comparison, and their architectures are illustrated in Table \ref{tb:cnn na} \citep{lecun2015deep}. Here, 8$\times$8 Conv.
32 represents a convolutional layer with 32 filters of size
8$\times$8, and each convolutional layer is followed by a batch normalization (BN) layer \citep{ioffe2015batch, agarap2018deep}.

\begin{table}[h]
	\centering
	\caption{The CNN-based network architectures.}
	\label{tb:cnn na}
	\begin{tabular}{llll}
		\hline
		Module & Policy network                                                                                                                                                                                        & Value network                                                                                                                                                & Encoder                                                                                                                                                        \\ \hline
		Input  & States                                                                                                                                                                                                & States                                                                                                                                                       & States                                                                                                                                                         \\ \hline
		Arch.  & \begin{tabular}[c]{@{}l@{}}8$\times$8 Conv 32, ReLU\\ 4$\times$4 Conv 64, ReLU\\ 3$\times$3 Conv 32, ReLU\\ Flatten\\ Dense 512, ReLU\\ Dense $|\mathcal{A}|$\\ Categorical Distribution\end{tabular} & \begin{tabular}[c]{@{}l@{}}8$\times$8 Conv 32, ReLU\\ 4$\times$4 Conv 64, ReLU\\ 3$\times$3 Conv 32, ReLU\\ Flatten\\ Dense 512, ReLU\\ Dense 1\end{tabular} & \begin{tabular}[c]{@{}l@{}}8$\times$8 Conv 32, ReLU\\ 4$\times$4 Conv 64, ReLU\\ 3$\times$3 Conv 32, ReLU\\ Flatten\\ Dense 512, ReLU\\ Dense $d$\end{tabular} \\ \hline
		Output & Actions                                                                                                                                                                                               & Predicted values                                                                                                                                             & Encoding vectors                                                                                                                                               \\ \hline
	\end{tabular}
\end{table}

For each game, we trained the agent for 10 million environment steps, in which the agent was set to interact with 10 parallel environments. Take PPO for instance, the agent sampled 256 steps in each episode, producing 2560 pieces of transitions. After that, the transitions were used to update the policy network and value network. The agent was trained with a learning rate of 0.0003, an action entropy coefficient of 0.01, a value function coefficient of 0.5, a generalized-advantage-estimation (GAE) parameter of 0.95, and a gradient clipping threshold of 0.1 \citep{schulman2015high}. More detailed parameters of PPO and A2C can be found in Table~\ref{tb:ppo atari training parameter} and Table~\ref{tb:a2c atari training parameter}.

\begin{table}[h]
	\caption{Hyparameters of PPO+REVD for Atari experiments.}
	\label{tb:ppo atari training parameter}
	\begin{center}
		\begin{tabular}{lll}
			\hline
			Method                                    & Hyperparameter             & Value   \\ \hline
			\multirow{15}{*}{PPO}                    
			& Observation downsampling   & (84, 84) \\ 
			& Stacked frames             & 4        \\
			& Environment steps          & 10000000 \\
			& Number of workers          & 10     \\
			& Episode steps              & 256     \\
			& Optimizer                  & Adam    \\
			& Learning rate              & 0.0003 \\
			& GAE coefficient            & 0.95    \\
			& Action entropy coefficient & 0.05    \\
			& Value loss coefficient     & 0.5     \\
			& Max gradient norm          & 0.5     \\
			& Value clipping coefficient & 0.2     \\
			& Batch size       & 64      \\
			& Update epochs              & 5       \\
			& Gamma $\gamma$             & 0.99    \\ \hline
			\multicolumn{1}{c}{\multirow{5}{*}{REVD}} & Embedding size $d$         & 128      \\
			\multicolumn{1}{c}{}                      & $k$                        & 5       \\
			\multicolumn{1}{c}{}                      & $\alpha$                        & 0.5       \\
			\multicolumn{1}{c}{}                      & $\lambda_{0}$              & 0.1     \\
			\multicolumn{1}{c}{}                      & $\kappa$                   & 0.00001 \\
			\multicolumn{1}{c}{}                      & $\epsilon$                 & 0.0001  \\ \hline
		\end{tabular}
	\end{center}
\end{table}

\begin{table}[h]
	\caption{Hyparameters of A2C+REVD for Atari experiments.}
	\label{tb:a2c atari training parameter}
	\begin{center}
		\begin{tabular}{lll}
			\hline
			Method                                    & Hyperparameter             & Value   \\ \hline
			\multirow{14}{*}{A2C}                     
			& Observation downsampling   & (84, 84) \\ 
			& Stacked frames             & 4        \\
			& Environment steps          & 10000000 \\
			& Number of workers          & 10     \\
			& Episode steps              & 32     \\
			& Optimizer                  & Adam    \\
			& Learning rate              & 0.0003 \\
			& GAE coefficient            & 0.95    \\
			& Action entropy coefficient & 0.01    \\
			& Value loss coefficient     & 0.5     \\
			& Max gradient norm          & 0.5     \\
			& Value clipping coefficient & 0.2     \\
			& Batch size                 & 32      \\
			& Gamma $\gamma$             & 0.99    \\ \hline
			\multicolumn{1}{c}{\multirow{5}{*}{REVD}} & Embedding size $d$         & 128      \\
			\multicolumn{1}{c}{}                      & $k$                        & 5       \\
			\multicolumn{1}{c}{}                      & $\alpha$                        & 0.5       \\
			\multicolumn{1}{c}{}                      & $\lambda_{0}$              & 0.1     \\
			\multicolumn{1}{c}{}                      & $\kappa$                   & 0.00001 \\
			\multicolumn{1}{c}{}                      & $\epsilon$                 & 0.0001  \\ \hline
		\end{tabular}
	\end{center}
\end{table}

\subsection{Experimental Setting for Exploration Methods}
\textbf{REVD.} We employed a randomly-initialized and fixed encoder for encoding the state space, whose architecture is illustrated in Table \ref{tb:cnn na}. At the end of each episode, the transitions were used to compute the mixed rewards using Eq.~(\ref{eq:total reward}), where $k=5, \alpha=0.5,\lambda_{0}=0.1,\kappa=0.00001$ and $\epsilon=0.0001$. In particular, since we considered the on-policy setting, the $k$-NN searching operation for a state was only performed within its own trajectory.

\textbf{RE3.} We followed the implementation of RE3 in a publicly released repository (\url{https://github.com/younggyoseo/RE3}), which uses intrinsic reward $\hat{r}(\bm{s}_t)=\mu_{k}(\bm{e}_{t}, \mathcal{E})$ without log exploration. Then the intrinsic reward was combined with the extrinsic reward to make a mixed reward $r^{\rm total}_{t}=\check{r}(\bm{s}_{t},\bm{a}_{t})+\lambda_{t}\cdot\hat{r}(\bm{s}_{t})$, where $\lambda_{t}=\lambda_{0}(1-\kappa)^{t}$. Here, we set $\lambda=0.05$ and $\kappa\in\{0.001, 0.0001, 0.00001\}$ and $k\in\{5, 10\}$. Similar to REVD, the $k$-NN searching operation for a state was also performed within its own trajectory.

\textbf{RIDE.} We follow the implementation of RIDE in a publicly released repository (\url{https://github.com/facebookresearch/impact-driven-exploration}). In practice, we trained a single forward dynamics model $g$ to predict the encoded next-state $\phi(\bm{s}_{t+1})$ based on the current encoded state and action $(\phi(\bm{s}_{t}),\bm{a}_{t})$, whose loss function is $\Vert g(\phi(\bm{s}_{t}),\bm{a}_{t})-\phi(\bm{s}_{t+1})\Vert_{2}$. To compute the state visitation frequency of $\bm{s}_{t+1}$, we used a pseudo-count method that approximates the frequency using the $k$-NN distance within episode \citep{badia2020never}.

\section{Details on PyBullet Experiments}\label{appendix:ddcse}

\subsection{Environment Setting}
We further tested REVD on six tasks from PyBullet Robotics Environments, namely \textit{Ant}, \textit{Cart Pole}, \textit{Half Cheetah}, \textit{Hopper}, \textit{Humanoid}, and \textit{Walker 2D}. The source code of these environments can be found in (\url{https://github.com/bulletphysics/bullet3}).

\subsection{Experimental Setting for RL}
Since PyBullet tasks provided low-dimensional features as observations, multilayer perceptron (MLP) \citep{lecun2015deep} was used to build the policy network and value network, whose architectures are illustrated in Table \ref{tb:mlp na}.  

\begin{table}[h]
	\caption{The MLP-based network architectures.}
	\label{tb:mlp na}
	\begin{center}
		\begin{tabular}{llll}
			\hline
			\textbf{Module} & Policy network                                                                                         & Value network                                                                     & Encoder                                                                           \\ \hline
			Input  & States                                                                                                 & States                                                                            & States                                                                            \\ \hline
			Arch.  & \begin{tabular}[c]{@{}l@{}}Dense 64, Tanh\\ Dense 64, Tanh\\ Dense dim($\mathcal{A}$)\\ Gauss Distribution\end{tabular} & \begin{tabular}[c]{@{}l@{}}Dense 64, Tanh\\ Dense 64, Tanh\\ Dense 1\end{tabular} & \begin{tabular}[c]{@{}l@{}}Dense 64, ReLU\\ Dense 64, ReLU\\ Dense $d$\end{tabular} \\ \hline
			Output & Actions                                                                                                & Predicted values                                                                  & Encoding vectors                                                                  \\ \hline
		\end{tabular}
	\end{center}
\end{table}

For each PyBullet task, we trained the agent for 2 million environment steps, in which the agent was also set to interact with 10 parallel environments. Take PPO for instance, the agent sampled 128 steps in each episode, producing 1280 pieces of transitions. After that, the transitions were used to update the policy network and value network. The agent was trained with a learning rate of 0.0003, an action entropy coefficient of 0.01, a value function coefficient of 0.5, a generalized-advantage-estimation (GAE) parameter of 0.95, and a gradient clipping threshold of 0.1 \citep{schulman2015high}. More detailed parameters of PPO and A2C can be found in Table \ref{tb:ppo bullet training parameter} and Table \ref{tb:a2c bullet training parameter}.

\begin{table}[h]
	\caption{Hyparameters of PPO+REVD for PyBullet experiments.}
	\label{tb:ppo bullet training parameter}
	\begin{center}
\begin{tabular}{lll}
	\hline
	Method                                    & Hyperparameter             & Value   \\ \hline
	\multirow{13}{*}{PPO}                     & Environment steps          & 2000000 \\
	& Number of workers          & 10     \\
	& Episode steps              & 128     \\
	& Optimizer                  & Adam    \\
	& Learning rate              & 0.0003 \\
	& GAE coefficient            & 0.95    \\
	& Action entropy coefficient & 0.01    \\
	& Value loss coefficient     & 0.5     \\
	& Max gradient norm          & 0.5     \\
	& Value clipping coefficient & 0.2     \\
	& Batch size                 & 64      \\
	& Update epochs              & 5       \\
	& Gamma $\gamma$             & 0.99    \\ \hline
	\multicolumn{1}{c}{\multirow{6}{*}{REVD}} & Embedding size $d$         & 64      \\
	\multicolumn{1}{c}{}                      & $k$                        & 3       \\
	\multicolumn{1}{c}{}                      & $\alpha$                        & 0.5       \\
	\multicolumn{1}{c}{}                      & $\lambda_{0}$              & 0.1     \\
	\multicolumn{1}{c}{}                      & $\kappa$                   & 0.00001 \\
	\multicolumn{1}{c}{}                      & $\epsilon$                 & 0.0001  \\ \hline
\end{tabular}
	\end{center}
\end{table}

\begin{table}[h]
	\caption{Hyparameters of A2C+REVD for PyBullet experiments.}
	\label{tb:a2c bullet training parameter}
	\begin{center}
		\begin{tabular}{lll}
			\hline
			Method                                    & Hyperparameter             & Value   \\ \hline
			\multirow{12}{*}{A2C}                     & Environment steps          & 2000000 \\
			& Number of workers          & 10     \\
			& Episode steps              & 8     \\
			& Optimizer                  & Adam    \\
			& Learning rate              & 0.0003 \\
			& GAE coefficient            & 0.95    \\
			& Action entropy coefficient & 0.01    \\
			& Value loss coefficient     & 0.5     \\
			& Max gradient norm          & 0.5     \\
			& Value clipping coefficient & 0.2     \\
			& Batch size                 & 32      \\
			& Gamma $\gamma$             & 0.99    \\ \hline
			\multicolumn{1}{c}{\multirow{6}{*}{REVD}} & Embedding size $d$         & 64      \\
			\multicolumn{1}{c}{}                      & $k$                        & 3       \\
			\multicolumn{1}{c}{}                      & $\alpha$                        & 0.5       \\
			\multicolumn{1}{c}{}                      & $\lambda_{0}$              & 0.1     \\
			\multicolumn{1}{c}{}                      & $\kappa$                   & 0.00001 \\
			\multicolumn{1}{c}{}                      & $\epsilon$                 & 0.0001  \\ \hline
		\end{tabular}
	\end{center}
\end{table}

\subsection{Experimental Setting for Exploration Methods}

For REVD, We also used a randomly-initialized and fixed encoder to encode the state space, whose architecture is shown in Table \ref{tb:mlp na}. In each episode, the visited states were first encoded as vectors with a dimension of 64. Then the encoding vectors were used to compute intrinsic rewards, where $k=3, \alpha=0.5, \lambda_{0}=0.1, \kappa=0.00001$, and $\epsilon=0.0001$. Finally, the augmentated transitions were used to update the policy network and value network. For RE3 and RIDE, we followed the similar procedures elaborated in Atari experiments and reported the best result.

\end{document}